\documentclass[]{interact}

\usepackage{epstopdf}
\usepackage{subfigure}
\usepackage[flushleft]{threeparttable}
\usepackage{multirow}

\usepackage{natbib}
\bibpunct[, ]{(}{)}{,}{a}{}{,}
\renewcommand\bibfont{\fontsize{10}{12}\selectfont}

\theoremstyle{plain}

\theoremstyle{definition}

\theoremstyle{remark}

\begin{document}

\articletype{RESEARCH ARTICLE}

\title{Automatic extraction of road intersection points from USGS historical map series using deep convolutional neural networks}

\author{
\name{Mahmoud Saeedimoghaddam\textsuperscript{a} and T.~F. Stepinski\textsuperscript{b}\thanks{CONTACT Mahmoud Saeedimoghaddam. Email: kiasaeedi.m@gmail.com; T.~F. Stepinski. Email: stepintz@uc.edu}}
\affil{\textsuperscript{a}Geography \& GIS department, University of Cincinnati, Cincinnati, OH, USA; \textsuperscript{b}Space Informatics Lab, University of Cincinnati, Cincinnati, OH, USA}
}

\maketitle

\begin{abstract}
	Road intersections data have been used across different geospatial applications and analysis. The road network datasets dating from pre-GIS years are only available in the form of historical printed maps. Before they can be analyzed by a GIS software, they need to be scanned and transformed into the usable vector-based format. Due to the great bulk of scanned historical maps, automated methods of transforming them into digital datasets need to be employed. Frequently, this process is based on computer vision algorithms. However, low conversion accuracy for low quality and visually complex maps and setting optimal parameters are the two challenges of using those algorithms. In this paper, we employed the standard paradigm of using deep convolutional neural network for object detection task named region-based CNN for automatically identifying road intersections in scanned historical USGS maps of several US. cities. We have found that the algorithm showed higher conversion accuracy for the double line cartographic representations of the road maps than the single line ones. Also, compared to the majority of traditional computer vision algorithms RCNN provides more accurate extraction. Finally, the results show that the amount of errors in the detection outputs is sensitive to complexity and blurriness of the maps as well as the number of distinct RGB combinations within them. 
	
\end{abstract}

\begin{keywords}
Road network data; object detection; GIS; Faster RCNN; deep learning
\end{keywords}

\section{Introduction}

Road networks data, and, in particular, road intersections data, have been used across different geospatial applications and analysis \citep{Chiang2009}. The intersections, treated as point features, have been used for georeferencing raster maps and for aligning geospatial datasets \citep{Chiang2005,Chen2008}. In addition, road intersection data were utilized to extract whole road networks from raster maps \citep{Chiang2009}. More recently, road intersection point data have been used in the fractal analysis of urban sprawl \citep{Murcio2015} and for obtaining the natural boundaries of cities in an objective way \citep{Masucci2015,Long2016}. Since the advent of the geographic information system (GIS), a large number of road network datasets become available in a vector format. However, road network datasets dating from pre-GIS years are only available in the form of printed maps and need to be scanned and processed before they can be analyzed by a GIS software \citep{Chiang2013}. This represents a significant barrier to their analysis.
Thus, scanning the data from physical maps and transforming them into the usable vector-based format is a prerequisite to temporal analysis based on road intersections. Due to the bulk of historical data, automated methods of transforming physical maps into digital datasets need to be employed.

Frequently, such data transformation process is based on computer vision algorithms \citep{Chiang2005,Chiang2009,Henderson2009,Henderson2014}, however setting optimal parameters in such algorithms requires experience, and, therefore, using them is not an option for users who are non-experts in the field of computer vision \citep{Ball2017,Uhl2018}. Moreover, due to low quality of many historical maps and the high rate of overlap of graphical features, the accuracy of the conversion is often low \citep{Chiang2009,Pezeshk2011}.

Using supervised machine learning algorithms instead of computer vision algorithms for the map conversion task can improve the accuracy, and it is more accessible to non-experts. In particular, a group of supervised machine learning algorithms called deep Convolutional Neural Networks (CNN) is a very good fit for the task of automated extraction of road intersections from physical maps. Deep CNN automatically selects the best attributes for distinguishing between intersections and other objects on the map, thus using it does not require user judgment on attribute selection \citep{Ball2017}.
In recent years deep CNN algorithms have been thoroughly studied and applied to the task of object recognition in an image. For example, deep CNN has increasingly been utilized in remote sensing tasks including geospatial object extraction from Earth observation data \citep{Wang2015,Tao2016,Amit2017,Ding2018}. As a result of the high interest in these algorithms, they have attained a high level of performance and are broadly recognized as the best choice for such applications \citep{Wu2017,Akcay2018,Lu2018}.

Identifying road intersections in scanned maps (images) is an example of an object recognition task, therefore it can be expected that deep CNN will perform well when applied to such a problem. Several existing studies used deep CNN for extracting geospatial features from historical maps (e.g. human settlements \citep{Uhl2017} and railroads \citep{Duan2018}), however, to the best of our knowledge, none of the previous deep CNN works have addressed the problem of road intersection extraction from physical maps.

In this paper, we employ the standard paradigm of using deep CNN for object detection task \citep{Jiang2017} named region-based CNN (RCNN) \citep{Girshick2014} for the task of identifying road intersections in scanned historical USGS (United States Geological Survey) maps of several US. cities. We extracted intersections from both single line and double line cartographic representations of the road maps. 

The remainder of this paper is organized as follows. Section 2 explains the general architecture of deep CNN and of the RCNN framework as well as its version used to identify road intersections. Section 3 describes the data and data capturing and preparation process as well as the parameters and indices selected for implementing, running and validating the intersection-detection algorithm. In section 4 we present the results and discuss their accuracy. Finally, the conclusion and future research directions are presented in Section 5.

\section{Method}
In this section, deep CNN and its design are explained and also the general idea of an object detection problem is outlined. Then, the way that deep CNN addresses object detection tasks is discussed by introducing the standard paradigm of deep CNN based object detection tasks named RCNN. Finally, the latest version of RCNN named Faster RCNN framework which has been used in this study is described.  
\subsection{Deep CNN}
Deep learning is a subdivision of supervised machine learning which is mainly referred to deep neural networks. The term 'deep' points to the idea of consecutive layers of data representations. Deep neural networks contain tens or hundreds of successive layers which produce meaningful representations of the data increasingly \citep{Abrishami2018}, while other approaches to neural networks contain only a few layers of representations \citep{Zarbaf2018}. In other words, through a multistage structure, deep neural networks distill the data and make a meaningful compact representation of them which is useful for a certain task \citep{Chollet2017}. For the computer vision tasks like image classification and object detection, a particular type of deep neural networks called deep CNN is predominantly used. 

The structure of deep CNNs consists of two major parts with different layers of data representations: 1- Feature extraction and 2- Classification. There are three types of layers in feature extraction part including convolutional layers, Rectified Linear Unit (RELU) layers, pooling layers (usually maximum pooling layer). Also, one or several fully-connected (fully connected neural network) layers exist in classification part \citep{OShea2015}. A convolutional layer extracts various shapes or features (e.g. horizontal line) from the input image by the moving kernels of various filters. Because convolution is a linear operation, a RELU layer adds the nonlinearity of real world data to the network by replacing the negative pixels within the output of the convolutional layer by zero. The final result of a convolution-RELU combination is called a feature map. A max pooling layer down-samples the feature maps by a maximum filter. It obtains a compact feature representation which remains steady to middling changes in the shape of the objects \citep{Goodfellow-et-al-2016}. A deep CNN architecture has several convolutional, RELU and pooling layers in a feedforward network which gradually extract more complex (high level) features from the image and produce a compact representation of them at the end. The obtained high level features of the image are then fed into the fully connected layers which classify the input images into a defined number of classes. Actually, a fully connected layer looks at the output of the last max pooling layer and determines which features highly correlate to a specific class \citep{Khan2018}. The training phase of a deep CNN consists of determining the best filters for the convolutional layers that best describe the features of the input images as well as adjusting the weights and biases of the fully connected layers. These are conducted by reducing the value of the loss function by gradient descent algorithm each time a new training image is fed into the network (back-propagation). Different numbers and orders of the layers result in different designs of deep CNNs such as AlexNet \citep{Krizhevsky2012}, Inception \citep{Szegedy2015}, ResNet \citep{He2016} etc.

\subsection{Object detection problem, RCNN, and Faster RCNN}
Although deep CNNs have been mainly used in image classification tasks \citep{Cetinic2018,Yang2018}, they can also be utilized in the object detection projects. The goal in an object detection problem consists of two tasks: First, localizing the parts of the image which contain predefined target objects; Second, precisely fitting the bounding boxes around the detected objects \citep{Naik2018}. Using a moving window over a set of images with ground truth boxes we can train a deep CNN for classifying the parts with the target objects and the parts without them (background). Because there is no guarantee that the sliding window fits the objects precisely \citep{Dickerson2017}, the deep CNN must be trained for refining the boundaries of the window using a bounding box regression model. The regression model is a neural network which is trained to transform and map the windows to the ground truth boxes by learning the adjustment factors \citep{Dickerson2017}. Searching the entire image by a moving window is computationally expensive, and thus many algorithms have been invented to optimize this process \citep{Alexe2012,Uijlings2013,Zitnick2014}. The main idea of the region proposal algorithms is instead of classifying all the possible windows on the image by a deep CNN, it is more efficient to find and propose the regions of the image which are the most promising candidates and allocate the computational budget to them \citep{Krahenbuhl2015}.  

Recently, many deep CNN-based object detection frameworks have been presented with outstanding performances some of which are Single Shot MultiBox Detector (SSD) \citep{Liu2016}, You only look once (YOLO) \citep{Redmon2017} and Region-based CNN (RCNN) \citep{Girshick2014}. The latter is the most successful object detector \citep{Deng2017} which proposes the potential regions using selective search algorithm \citep{Uijlings2013}. Then, it feeds the proposed regions into a deep CNN in order to extract their distinct features and classifying them as target/background. Finally, the bounding boxes of the target regions are tightened and refined using a trained regression model \citep{Girshick2014}. RCNN is a slow system since 1- Selective search algorithm is applied within an external distinct part from the deep CNN which causes a computational bottleneck and 2- Each proposed region is fed into a deep CNN separately to extract the final feature maps which takes a long time. To overcome these challenges, Fast RCNN \citep{Girshick2015} and more recently Faster RCNN \citep{Ren2017} have been presented.

\begin{figure}
	\centering
	\subfigure{
		\resizebox*{\textwidth}{!}{\includegraphics{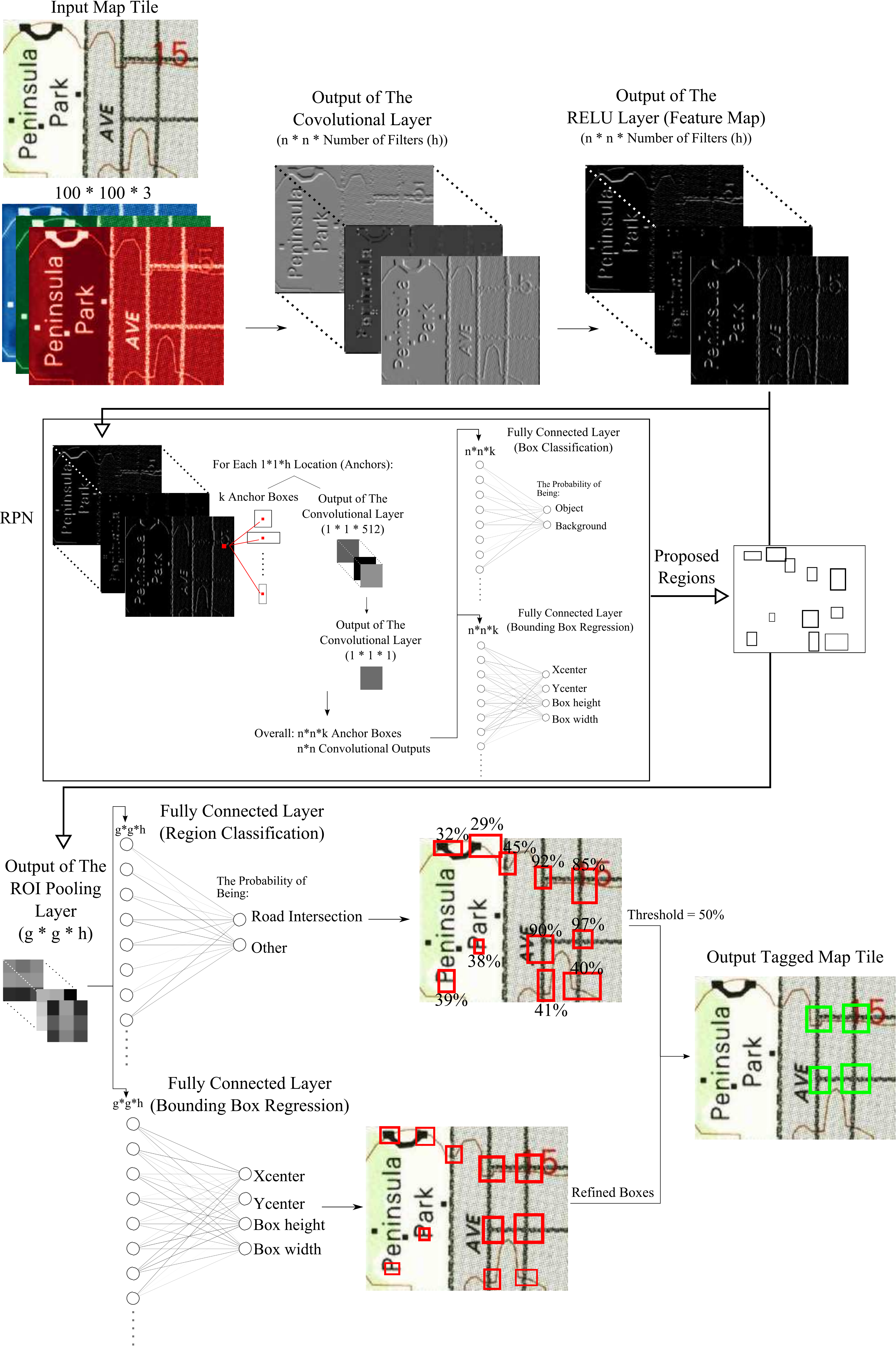}}}\hspace{20pt}
	\caption{The diagram of the Faster RCNN framework. During the training phase, the neurons related to the background boxes are disabled within the bounding box regression parts.} \label{fig1}
\end{figure}

In Faster RCNN framework (Figure \ref{fig1}), instead of using a separate selective search part to obtain the proposed regions from the input image, first the final feature maps (with the shape of $n\times n\times h$) of the image is extracted using a deep CNN and then in a Region Proposal Network (RPN) the candidates are obtained from the maps. Traditional methods like selective search are based on the segmentation algorithms which need feature extraction like edge detection \citep{Uijlings2013}; The final feature maps made by the deep CNN already contains all the features needed for region proposing so it is possible to reuse these maps and avoid the redundant computation. RPN slides a window on the final feature maps and for each location (anchor) it creates k boxes with different scales and aspect ratios (anchor boxes). The part of the feature maps within each window is fed into 2 consecutive convolutional layers which make $1\times1\times512$ and finally $1\times1\times1$ layers respectively for each one of the k anchor boxes ($n\times n\times k$ layers overall). The final layers are then fed into the two separate fully connected layers: One for classification to determine the probability of the boxes contain potential regions and the other for regression to refine the extents of the boxes. The output of the fully connected layers are the proposed candidate regions for further analysis. Note that during the training phase of the RPN, the anchor boxes are considered to be the potential regions if their overlaps with the ground truths are higher than a certain value (usually 0.7) and are considered to be backgrounds (without any potential objects) if the overlaps are less than a certain value (usually 0.1). Anchor boxes which are not qualified based on this range are not used in the training phase. Because the anchor boxes highly overlap each other, an algorithm named non-maximum suppression (NMS) is used to reduce the redundancy by defining the maximum number of proposals (usually 300).
In the next step of Faster RCNN framework, the feature maps within all proposed regions are extracted and a pooling layer named Region Of Interest (ROI) is applied to them. Then the result is fed into the two fully connected layers to obtain the probability of the proposed regions being the target objects (road intersections in our case) and to refine the bounding boxes of the regions. The loss function of the whole Faster RCNN framework is the sum of the loss values of the RPN and the final detection part \citep{Ren2017}

The loss value of the RPN is calculated by combining the boxes classification loss and the loss of the regression which refines their extents (Equation \ref{loss_RPN}):
\begin{equation} \label{loss_RPN}
L(p_{i}, t_{i}) = \frac{1}{N_{cls}} \sum_{i} L_{cls}(p_{i},p_{i}^*) + \lambda \frac{1}{N_{reg}} \sum_{i} p_{i}^* L_{reg}(t_{i},t_{i}^*) 
\end{equation} where $p_i$ is the predicted probability of anchor box \textit{i} being an object and $p_i ^ *$ is the ground truth label (1 for being an object and 0 for otherwise). $t_i$ is the coordinates (center coordinate of the box and the height and width of the it) of the predicted boundary of the anchor box \textit{i} and  $t_i ^ *$ is the coordinates of the ground-truth box. $L_{cls}$ is \textit{log} of the classification loss over two classes (object vs background) and $L_{reg}$ is the regression loss of refining the boxes. To avoid bias toward the background samples $L_{cls}$ is obtained for a batch of random anchor boxes with size $N_{cls}$. Also, $L_{reg}$ is calculated for the $N_{reg}$ boxes classified as an object. Finally, $\lambda$ is the hyper-parameter which controls the weights of the two parts in the loss function \citep{Ren2017}. 

The loss function of the final detection part of the system has a similar structure to Equation \ref{loss_RPN} in which $p_i$ is the predicted probability of the proposed region \textit{i} being the target object and $p_i ^ *$ is the related ground truth label. $t_i$ is the coordinates of the predicted boundary of the object \textit{i} and  $t_i ^ *$ is the coordinates of the ground-truth region. $L_{cls}$ is \textit{log} of the classification loss and $L_{reg}$ is the regression loss of refining the regions. $N_{cls}$ is the number of proposed regions to be classified and $N_{reg}$ is the number of regions classified as the target object \citep{Girshick2015}.

\section{Experiment}

\subsection{Data}
\begin{figure}
	\centering
	\subfigure{
		\resizebox*{\textwidth}{!}{\includegraphics{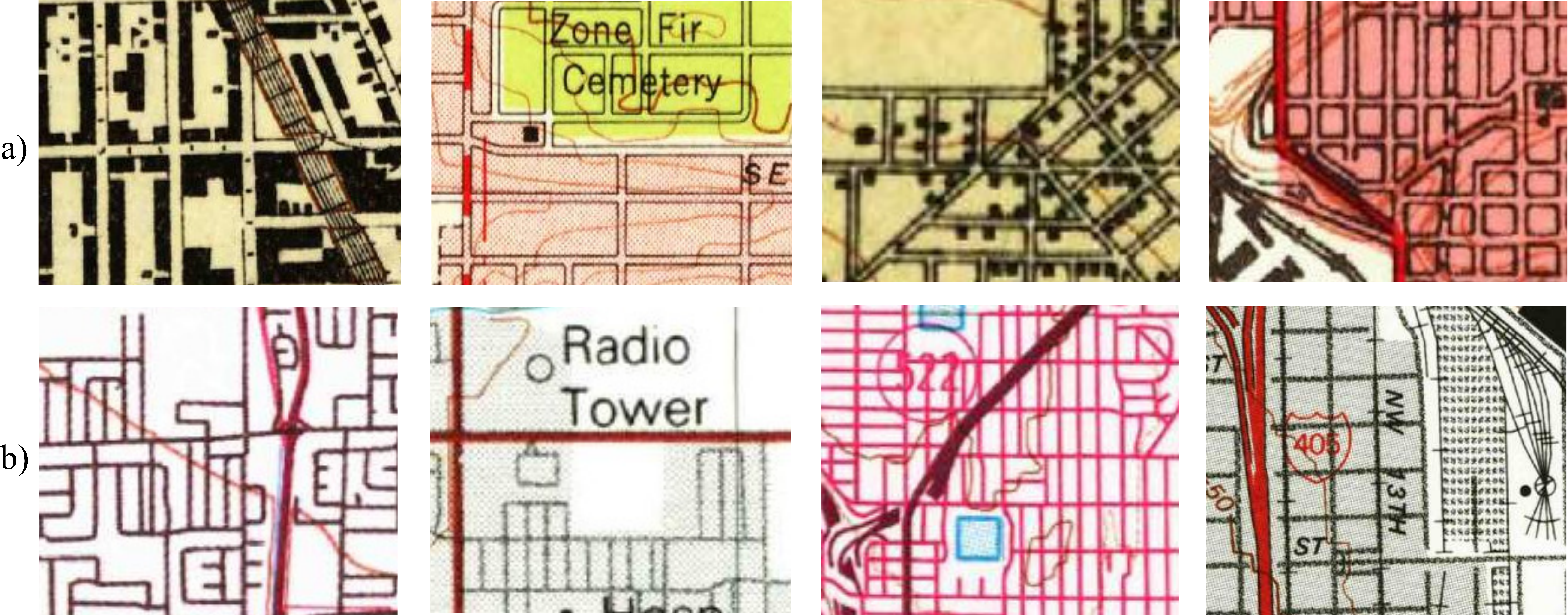}}}\hspace{20pt}
	\caption{Samples of map tiles with different cartographic style. a) Map tiles with double line symbol of light duty roads, b) Map tiles with single line symbol of light duty roads.} \label{fig2}
\end{figure}

From 1884 to 2006 USGS produced paper topographic maps. They were originally prepared for minerals exploration; however, they have been used in other applications gradually. Due to the growing needs of digital data for using in GIS, USGS decided to scan and georeferenced the map series in the mid-1990s. These digital raster graphics (DRGs) were mostly produced from 1995 to 1998 and about 1,000 new DRGs were added to the dataset during the next years \citep{Allord2014}. Now, the database contains more than 180,000 historical map sheets within the country \citep{Uhl2017}. In order to maintain the exact appearance of the hard copies of the maps in the scanned versions and also a high level of precision after georeferencing, USGS has determined 600 PPI (pixels per inch) as the resolution value. 8-bit depth was selected for the images as the proper value of precision with which colors are specified in them. There are different map scales of historical maps from 1: 24,000 to 1: 250,000 \citep{Allord2014}. Except for 1: 250,000 maps which do not contain all types of roads, all other maps with different scales have been used in our study. Same as the maps scales, standard symbols of some geographical entities have also changed during the 19 and 20 centuries \citep{USGS2005}. Although there are several symbols of different types of roads in the maps (highway, trail, etc.), most of the roads (especially in the urban areas) have been categorized as light duty roads. Thus, they play a vital role in the studies of road and road intersection detection. Consequently, we decided to categorize the maps into two sets (Figure \ref{fig2}): i) The maps with single line symbol of the light duty roads (Mostly in 1: 100,000 scales), ii) The maps with double line symbol of light duty roads. By this categorization, it is possible to compare the precision of the deep CNN in detecting the road intersections in the two cartographic styles with different complexity levels. 

\begin{figure}
	\centering
	\subfigure{
		\resizebox*{\textwidth}{!}{\includegraphics{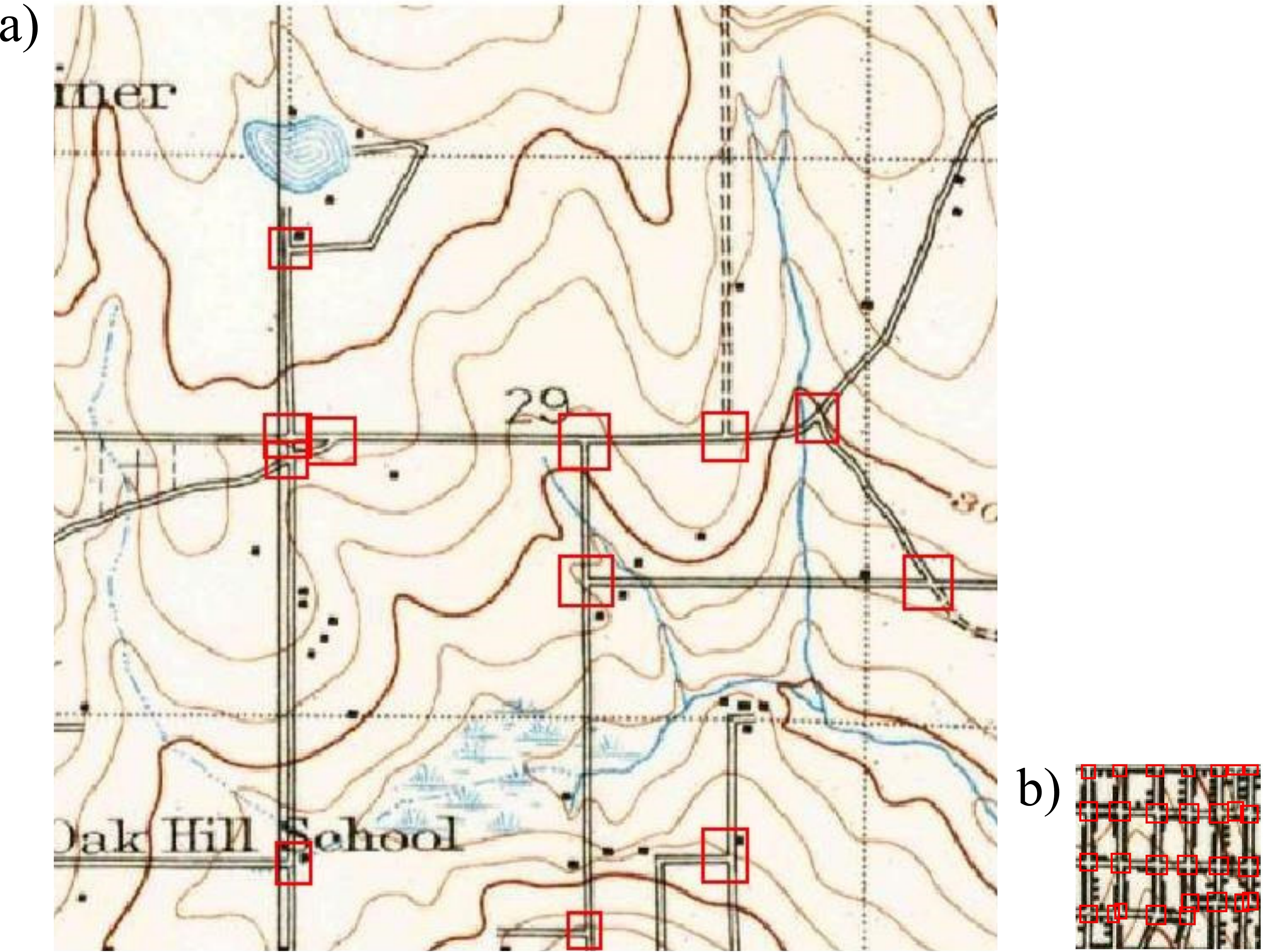}}}\hspace{20pt}
	\caption{Tagged map tiles (Seattle-WA-1908); a) 500*500 mostly negative sample with the small number of road intersections and more negative objects (contours, lake, annotations, etc.), b) 100*100 mostly positive sample with more road intersections.} \label{fig3}
\end{figure}

In order to make representative samples of the road intersections with different shapes and styles, we selected 23 cities in 15 states and captured the maps of different years (37 maps overall) (ngmdb.usgs.gov/topoview). We randomly selected tiles from these maps in the way that they include both positive and negative samples (Figure \ref{fig3}). In other words, selected map tiles consist of so many different road intersections as well as the parts with other entities (contour lines, rivers, annotations, boundaries, etc.) and background. The intersections in the tiles have been tagged manually. The sizes of the tiles with more road intersections (positive parts) were 100*100 pixels and the sizes of the tiles with more negative parts were 500*500 pixels. The bigger size of mostly negative tiles helped us to include different objects. Also, we used smaller tiles of positive samples for the sake of convenient and fast tagging process. About 12,000 road intersections have been tagged within 2,000 tiles of each map style (about 24,000 tags in 4,000 tiles overall). 

\begin{figure}
	\centering
	\subfigure{
		\resizebox*{\textwidth}{!}{\includegraphics{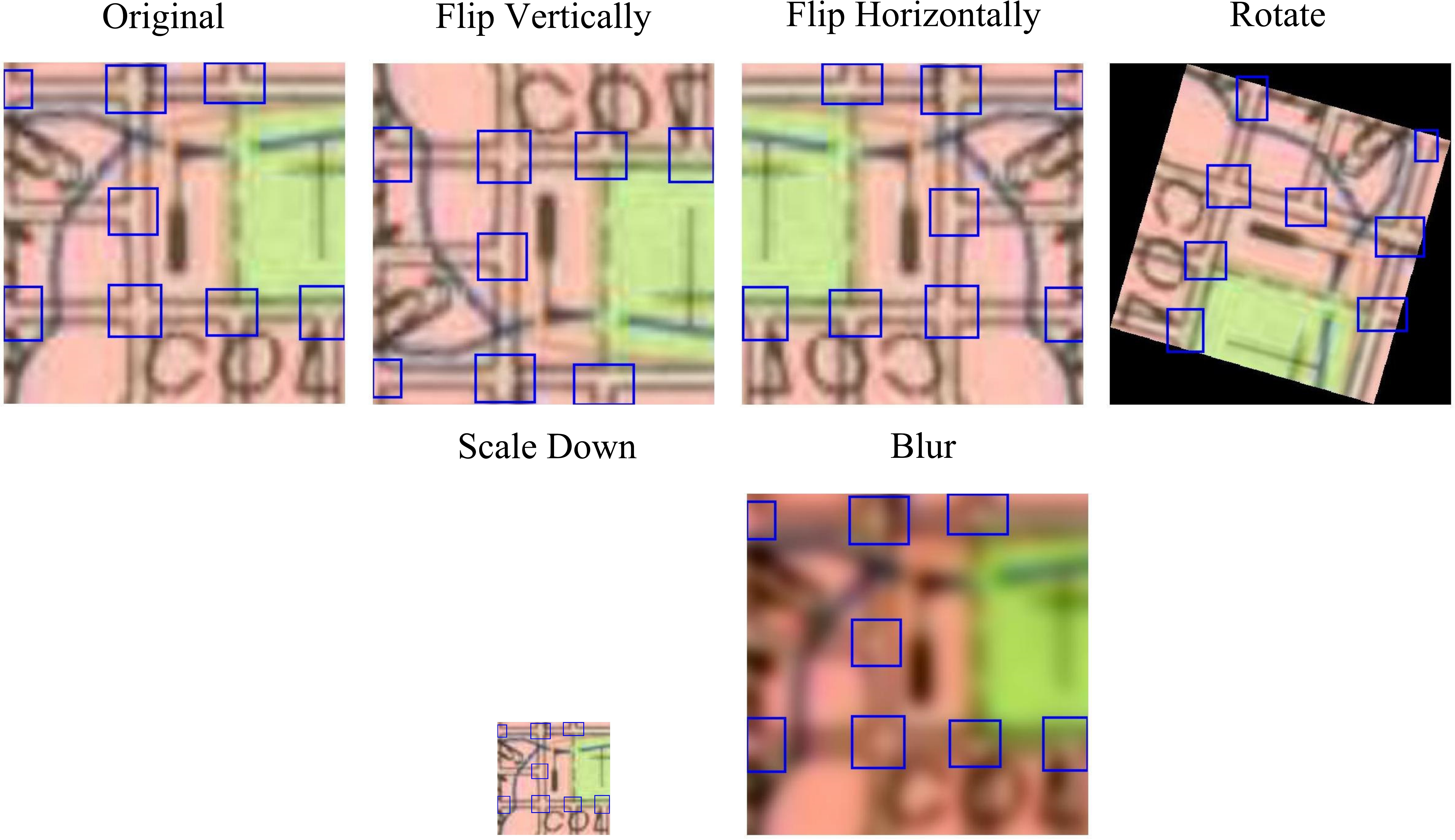}}}\hspace{20pt}
	\caption{Data augmentation using different methods} \label{fig4}
\end{figure}

Since data size is the key factor in the performance of deep CNNs \citep{Salamon2017} we enlarged the dataset to 10,000 tiles for each map style using a technique named data augmentation which has promoted the accuracy in the previous studies of object detection \citep{Jo2017,Lv2017,Xi2018}. We randomly used 5 types of augmentation methods (Figure \ref{fig4}): flip horizontally, flip vertically and rotate by a random degree in the range of ($0^\circ$, $360^\circ$) which provide different shapes and orientations of the intersection; Blur with Gaussian filter (random variance) and scale down by a random number between 3 and 5 which provide tiles with low graphical qualities.

\subsection{Parameter selection}

\subsubsection{Deep CNN architecture}
In the original paper, Faster RCNN utilized Zeiler and Fergus’s (ZF) \citep{Zeiler2014} and Simonyan and Zisserman’s (VGG) \citep{Simonyan2014} architectures for the deep CNN module. However, previous studies have shown that although Faster RCNN with ‘Inception Resnet v2’ multi-block architecture \citep{Szegedy2016} is quite slow, it attains the most accurate results \citep{Huang2017}. Thus, we decided to use this architecture in our system. Inception-ResNet v2 architecture combines the ideas of Inception or GoogleNet, which obtains greater number of meaningful objects at multiple sizes by stacking the outputs of convolutional and pooling layers with different moving window sizes and ResNet (residual network), which addresses the problem of demanding training phase of very deep networks and results in low loss values using a trick named 'skip connections' \citep{Rahnemoonfar2017,Ayrey2018}. In our study, the Faster RCNN framework used the output of "Inception-ResNet-B" block of the CNN architecture as the shared feature map. This feature map consists of 1088 filters which means it represents 1088 different high level features of the input image through a combination of 100 convolutional, 30 RELU and 3 max pooling layers. The fully connected layers regrading to box classification and regression of RPN module consist of $30,000\times 2$ weights plus 24 bias values and $30,000\times 4$ plus 48 bias values as the parameters respectively. Also, the fully connected layers regrading to box classification and regression of the final detection part include $1536\times 2$ weights plus 2 bias values and $1536\times 4$ weights plus 4 bias values respectively.  

Transfer learning is a technique in which low-level spatial features like edge and corner which are detected in the first layers of a network and are generic in all sources of images can be adapted to a new visual task with a new network \citep{Marmanis2016}. In other words, instead of initializing a set of random weights for the neural network to train from scratch we can easily readjust previously learned weights of another pre trained network. This process can save a lot of time (from several weeks to several hours) in the training phase. It has been previously shown that pre trained networks on large-scale image datasets like Microsoft COCO \citep{Lin2014} and ImageNet \citep{Deng2009} can be used in extracting geospatial objects from aerial images \citep{Marmanis2016}; Thus, In our study, we used the Inception ResNet v2 architecture pre trained on COCO dataset. This dataset consists of 2.5 million labeled tags in 328k images.

\subsubsection{RPN}
In the RPN part of our model (Model means the object proposal and classifier system) we selected 0.25, 0.5, 1.0, 2.0 as the scale values and 0.5, 1.0, 2.0 as the aspect ratio values to make 12 anchors at each sliding position. We defined the maximum value of proposals equal to 300 for NMS. 100 out of 300 proposed boxes with the highest probability of being an object were fed to the final detection part. Also, we used 0.7 and 0.1 as the thresholds of defining object and background boxes during the training phase of the RPN.

\subsubsection{Loss function}
We used the same weights for the classification and localization losses within the loss functions of both parts of the system. Also in adjusting the weights in each iteration of the neural networks, the technique named gradient descent is used. In gradient descent, we look for the minimum value of training loss by moving along the cost function step by step. Learning rate defines the length of these steps of movement. By specifying a small value for the learning rate it takes a long time to reach the minimum value. On the other hand, a large value of the rate might cause ignorance of the minimum. We used 0.003 for the rate which was defined in the pre trained model.  

\subsection{Model Evaluation}

For evaluating the models, we used 20\% of the tagged image (2,000 items) as the test data. The widely used evaluation indicators of object detection tasks are “Precision” and “Recall” which are calculated by Equation \ref{eval}a and \ref{eval}b \citep{Chen2018}: \begin{subequations}\label{eval}
\begin{equation}
Precision = \frac{True Positive}{True Positive+False Positive} 
\end{equation}\begin{equation}
Recall = \frac{True Positive}{True Positive+False Negative} 
\end{equation}
\end{subequations}Where True Positive (TP) is the number of correctly detected target objects and False positive (FP) is the number of objects that incorrectly detected as the target object. False negative (FN) is the number of target objects that are not detected by the model. Typically, these indices are estimated by an indicator named Intersection over Union (IOU). IOU is the percentage of overlap between a detected bounding region and the related ground truth. The detected region is considered as TP if the related IOU is above a certain percentage (usually 50\%) and FP otherwise. The ground truths without any corresponding detected regions are adjudged FN \citep{Everingham2010}. Since the extent of the bounding boxes of the road intersections are arbitrary and there is no predefined restriction and size to depict them on the map, the overlap between a ground truth and the corresponding detected region is not a suitable indicator. We decided to use another definition of TP, FP, and FN which is utilized by \cite{Chiang2009}; We considered the center of the detected regions as the road intersection points. Hence, a point which was in the corresponding ground truth was defined as TP and FP otherwise. Also, the ground truths without any points in them were considered as FN. It should be mentioned that if we had several points in a box only one of them was defined as TP. On the other hand, if we had a point which was located within several ground truths we recorded only one TP and the remaining regions were considered as FN.  

Finally, all the object detection tasks have been done using Google’s Tensorflow Object Detection API \citep{Huang2017} which is an open source framework built on top of TensorFlow package of python programming language that makes it convenient to train and deploy deep CNN based object detection models ($https://github.com/tensorflow/models/tree/master/research/object_detection$).

\section{Results and discussion}
In this section, first, the graphs of the loss functions are presented for each model (single line and double line styles) separately and compared. Then, the accuracies of the models are evaluated and collated with the previous methods. Also, some sample tiles of the detection results are presented as the tagged images in order to discuss them visually. The latter gives us the idea about the capability of the models to detect the intersections and also its errors in distinguishing them from other similar objects. Finally, the impacts of some characteristics of the map tile images on the detection accuracy are explored.

\subsection{The loss function}
\begin{figure}
	\centering
	\subfigure{
		\resizebox*{\textwidth}{!}{\includegraphics{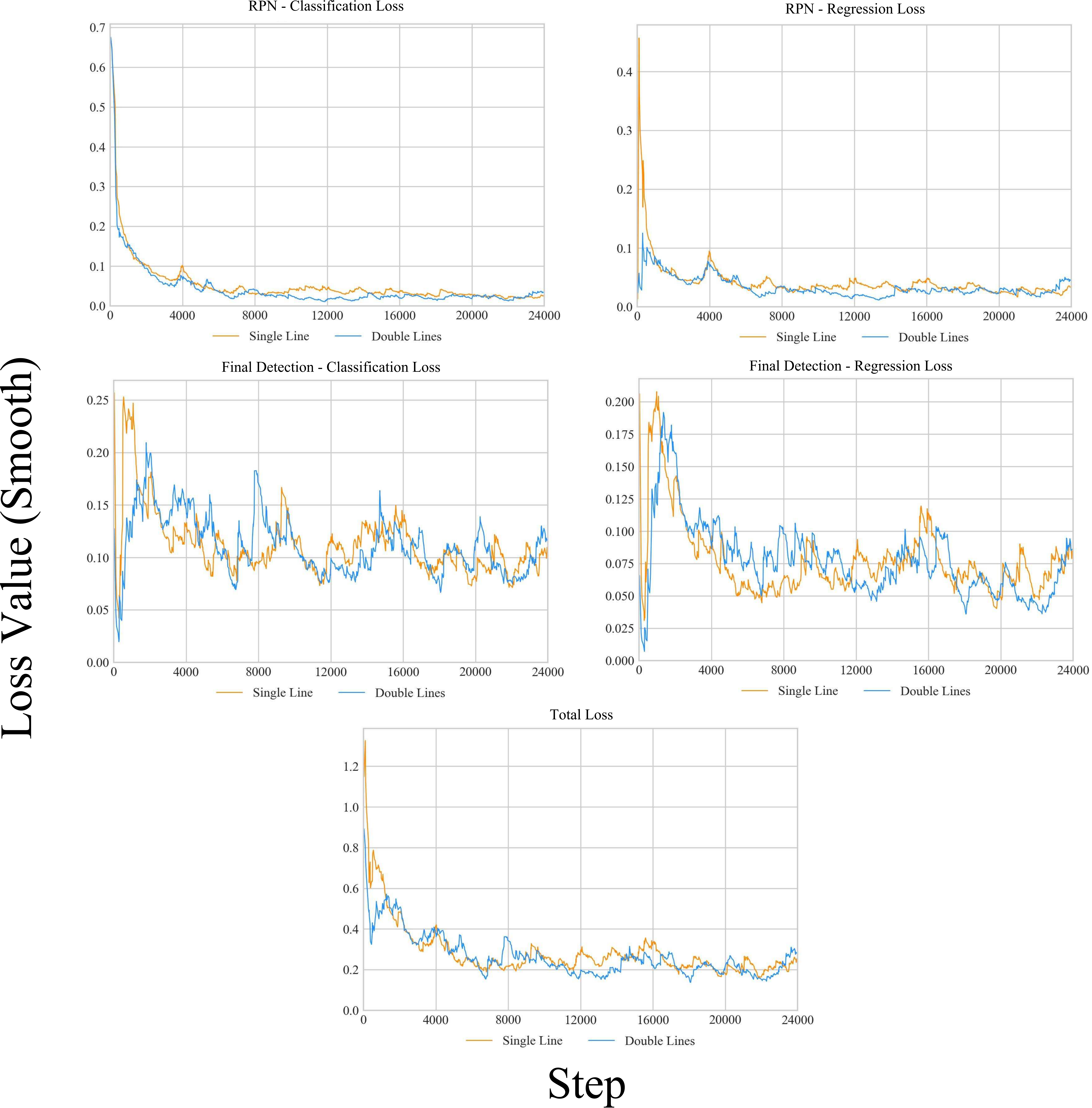}}}\hspace{20pt}
	\caption{The graphs of training loss related to the RPN, the final detection part and total after 24000 steps (3 epochs). The loss values were smoothed in order to reveal the convergence trend.} \label{fig5}
\end{figure}

Figure \ref{fig5} shows the total training loss of the system as well as the loss values of its RPN and the final detection parts regarding the two models after 24000 steps (3 epochs). In order to reduce the oscillations and reveal the convergence trends, the loss values have been smoothed. The total loss values of both models converge to 0.4. The loss value of the regression and the classification tasks of the RPN part converge to a number less than 0.1. The loss value of the final detection part oscillates around 0.3 and 0.07 in the classification and regression task respectively. All of the above convergences started after 12000 steps (1.5 epochs). Thus, both models need to go through the training data 1.5 times to adjust the weights of all parts of the system. Also, it can be noticed that both models show the same pattern during the training process. The fluctuations of the RPN loss value around the convergence line are less than the final detection part, especially for the regression task. It means that the gradient descent algorithm is able to find the minimum value of the cost function more conveniently in the RPN part compared with the final detection part. It makes sense because the task of the RPN part is just finding the potential regions in a tile which could be easier than the task of the final detection part (to discriminate the intersections). Finally, it can be noticed that in the majority of the graphs, loss value of the model of single line symbol (orange line) starts from a higher value and the range of its fluctuations is higher compared with the model of double line symbol (blue line). This fact shows that detecting the intersections from single line roads needs more weights modification and adjustment compared with double line roads. The resemblance of the single line road symbol to the other objects in the maps could be the main reason for this fact.

\subsection{Model accuracy}
All the detected boxes with 0.5 or more probability of containing an intersection were considered as the outputs of the models. Applying the method described in section 3.3, the precision and recall measures were calculated. Table \ref{table1} shows the accuracy of Faster RCNN framework and previous methods in extracting the road intersections from the USGS historical maps and compares them using F1-score which is the harmonic mean of precision and recall \citep{Chew2018}.

\begin{table}[h]
\caption{Accuracy of the deep learning method along with the previous methods in extracting the road intersections from the historical maps.}
\resizebox{\textwidth}{!}{
{\begin{tabular}{lccccc}\toprule
		Method & Algorithm & Road’s symbol type & Precision & Recall & F1-score \\ 
		\midrule
		\multirow{2}{*}{Deep learning} & \multirow{2}{*}{Faster RCNN} & Single line & 0.76 & 0.84 & 0.8\\ 
		& & Double line & 0.9 & 0.82 & 0.86\\
		\hline
		\multirow{4}{*}{Traditional computer vision} & \cite{Chiang2005} &
		 Double line & 0.84 & 0.75 & 0.79 \\
		 & \cite{Chiang2009} &
		 Double line & 0.82 & 0.6 & 0.69 \\
		 & \cite{Henderson2009} &
		 Single line & 0.66 & 0.93 & 0.77 \\ 
		 & \cite{Henderson2014} &
		 Single line & 0.8 & 0.82 & 0.81 \\ \bottomrule	
\end{tabular}}}
\begin{tablenotes}
	\footnotesize\item *Note that the map tiles used in all studies are from the same data source (USGS historical maps) but not necessarily identical. 
\end{tablenotes}

\label{table1}
\end{table}

\begin{figure}[h]
	\centering
	\subfigure{
		\resizebox*{\textwidth}{!}{\includegraphics{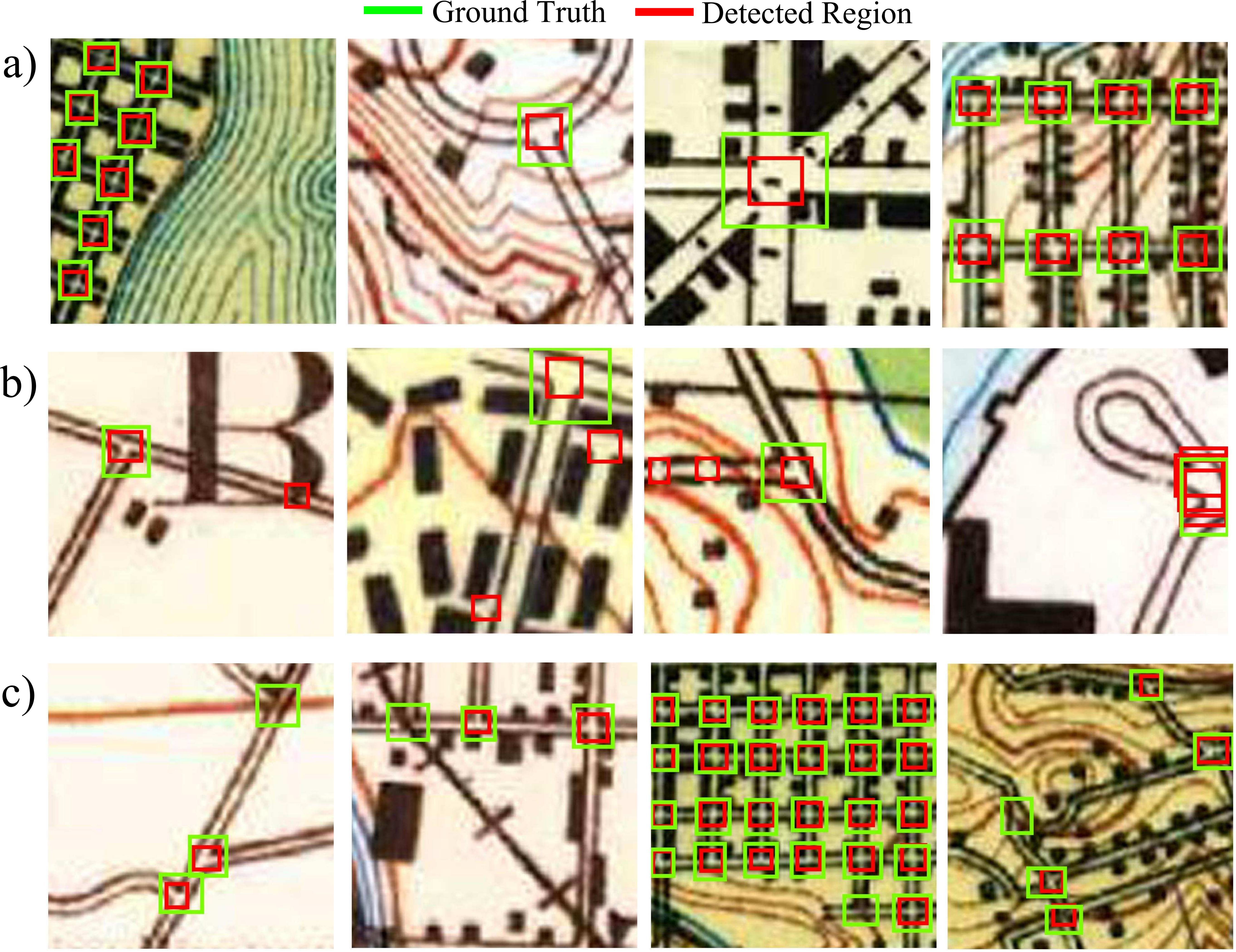}}}\hspace{20pt}
	\caption{Detection result of Faster RCNN framework for double line road symbol; a) Selected test tiles in which all the intersections have been detected, b) Selected test tiles contains FP errors, c) Selected test tiles contains FN errors.} \label{fig6}
\end{figure}

As it can be noticed from Table \ref{table1}, precision, recall, and F1-score of Faster RCNN framework are higher than the previous methods for the maps with double line symbols of the roads. Based on the precision value, it can be claimed that 90\% of the output boxes contain a road intersection averagely. Also, the recall value shows the fact that on average, the model is able to extract 82\% of the road intersections in a map. Figure \ref{fig6} shows the detection results in the selected test tiles with double line road symbol. In Figure \ref{fig6}-a all the intersections within the tiles of several cartographical styles and different complexity have been detected by the system. Figure \ref{fig6}-b shows the situations in which the system created FP boxes. The selected tiles of this part contain the typical examples of FP errors. FP boxes have been made where 1) A road is overlapped with other entities like annotations which makes a shape similar to a road intersection, 2) Some of the settlement objects which are close together creating a shape similar to a road intersection, 3) The gap exists in a dashed lines symbol of a trail road combined with another object like a contour making the shape of a road intersection and 4) Several boxes were proposed by RPN as different potential regions for a single anchor which makes a lot of boxes for a single road intersection. Finally, Figure \ref{fig6}-c shows the selected tiles that include FN boxes. As it can be noticed from the tiles, majority of FN errors happen where a linear object other than streets covers the intersection and significantly changes the structure of the road intersection and causes the system to misclassify the related proposed box.

\begin{figure}[h]
	\centering
	\subfigure{
		\resizebox*{\textwidth}{!}{\includegraphics{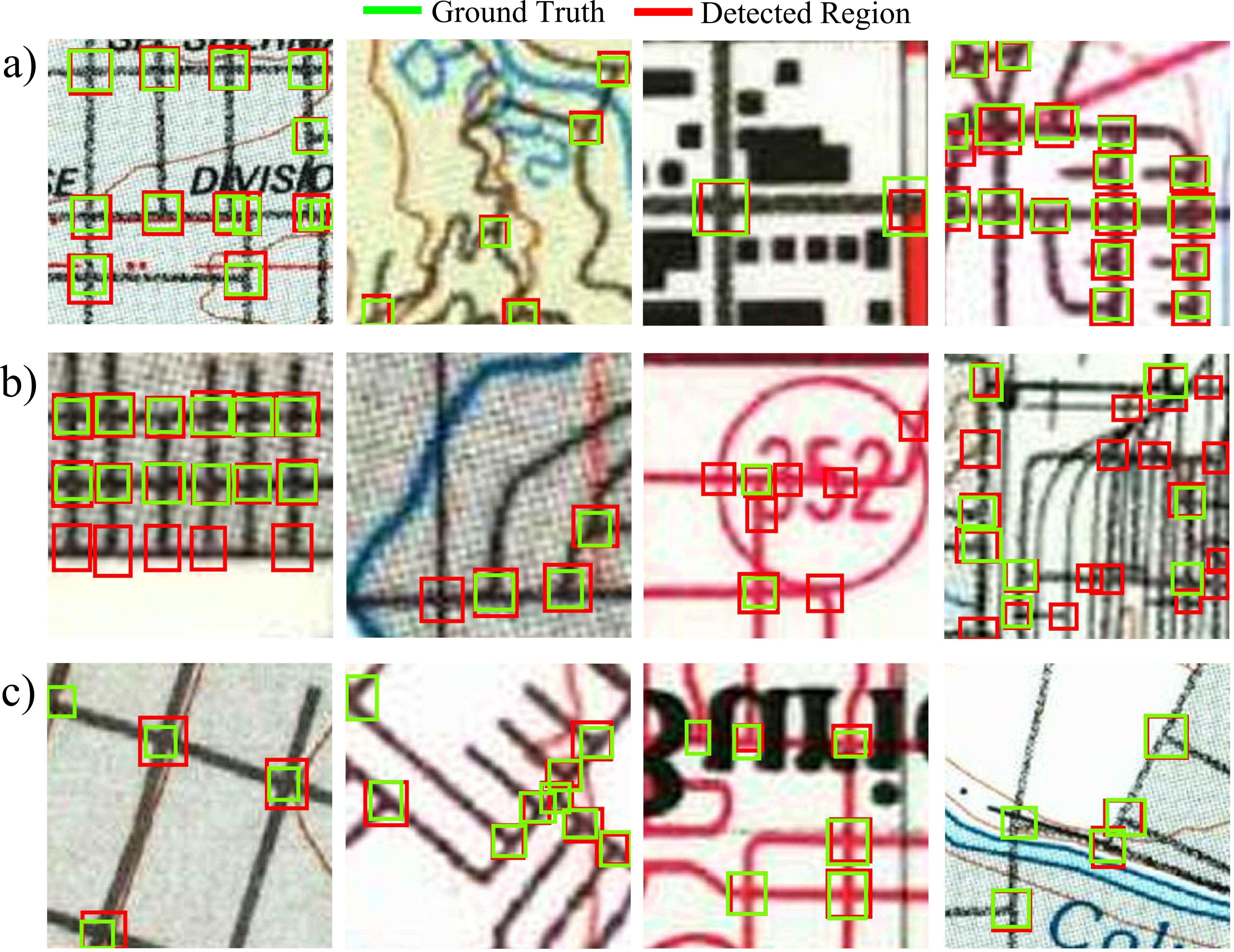}}}\hspace{20pt}
	\caption{Detection result of Faster RCNN framework for single line road symbol; a) Selected test tiles in which all the intersections have been detected, b) Selected test tiles contains FP errors, c) Selected test tiles contains FN errors.} \label{fig7}
\end{figure}

According to Table \ref{table1} it can be concluded that the precision of Faster RCNN framework in detecting the intersections is higher than the method of \cite{Henderson2009} and lower than the one described in \cite{Henderson2014} for the maps with single line symbols of the roads. On the other hand, the recall value of the deep learning method shows the converse analogy. Overall, Faster RCNN shows a higher F1-score than \cite{Henderson2009}'s method and lower than \cite{Henderson2014}'s. Based on the precision value, it can be claimed that 76\% of the detected boxes contain a road intersection averagely. In addition, the recall value shows the fact that, on average, the model is able to extract 84\% of the road intersections in a map. The overall accuracy of the model for single line symbol is lower than the one for double line symbol but its recall is higher. Figure \ref{fig7} demonstrates the detection result in the selected test tiles which contain single line road symbol with the same structure as Figure \ref{fig6}. Because of the resemblance between road lines and other linear objects like rivers, map grids, contours, etc. more FP errors have been made by the system (Figure \ref{fig7}-b). For instance, some parts of the map annotations might be classified as a road intersection. Also, the cross point of a road with the frame of the map might be attributed to a road intersection. In some cases, the similarity of the shape and color between a road line and the circumference of the routes signs makes the system to classify their crossing point as a road intersection. More importantly, recognizing the street intersections from railroad intersections is not as convenient as the maps with double line roads. Figure \ref{fig7}-c shows the selected tiles that include FN boxes. As it can be noticed from the tiles, FN errors might happen when the road intersection is too close to the boundaries of the map tile or another entity like an annotation. This might remove some parts of the shape of the intersection and makes it hard to be detected. 

\subsection{The effects of the characteristics of the map tiles on the accuracy}

\begin{figure}
	\centering
	\subfigure{
		\resizebox*{\textwidth}{!}{\includegraphics{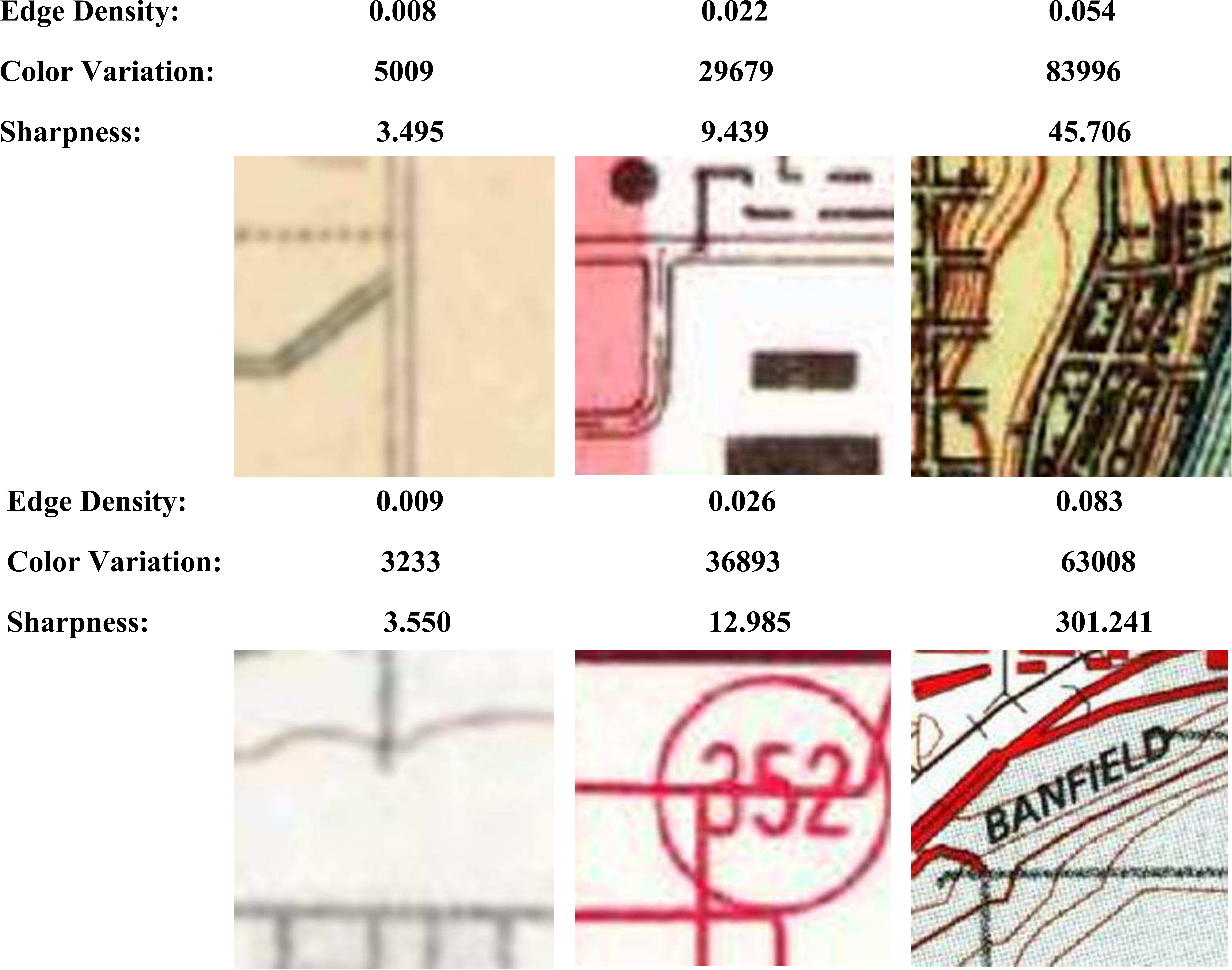}}}\hspace{20pt}
	\caption{The values of three different characteristics of the selected map tiles.} \label{fig8}
\end{figure}

In this part, the impacts of some characteristics of the map tile images on the detection accuracy are explored. We considered 3 types of characteristics including 1) Edge density which is the average of the edge magnitude (calculated by Sobel operator) in the entire image \citep{Phung2007}. Edges in an image defined as the parts where the brightness of the image changes significantly \citep{Gao2010}. Since the edges within an image play a vital role in shaping the objects, edge density could be a proper indicator of the clutter level in the image tiles and their complexity \citep{Bhanu1986}. A high degree of edge density causes highly textured images that present more of a challenge to an object detector framework \citep{Peters1990}; 2) The number of unique Red-Green-Blue (RGB) combinations in an image which shows the diversity of the input’s pixel values. 3) Sharpness (less blurriness) which was calculated using the variation of the image Laplacian \citep{Pech2000}. Figure \ref{fig8} shows the values of these metrics for 3 different map tiles.
As it can be seen in Figure \ref{fig8}, from left to right, the complexity and RGB variation increase because more edges and color codes have been added to the tiles. On the other hand, the level of blurriness decreases (the sharpness increases) which affected the quality of the images. Our goal in this part is to see if the high complexity, colorfulness, and blurriness of an image causes more FP and FN errors among 100 proposed boxes fed into the final detection part. This analysis can show us the sensitivity of the model accuracy to each one of the above variables. Thus, multiple linear regression was applied to investigate the possible effects. We used the standardized variables in order to compare the strengths of their impacts and address the multicollinearity problem. Table \ref{table2} shows the result of the regression analysis. 
\begin{table}[h]
	\tbl{The result of the multilinear regression analysis for evaluating the impact of image complexity, colorfulness and blurriness on the amounts of FP and FN errors. ($\alpha$ = 0.05)}
	{\begin{tabular}{lcccccc} \toprule
			\textbf{Dependent Variable: FP}\\
			\cmidrule{1-1}
			Double line symbol & Standardized coefficient & Standard error & t & P-value & 0.0250 & 0.9750 \\ \midrule
			Edge density & 0.370 & 0.064 & 5.767 & \textless 0.001 & 0.244 & 0.496 \\
			RGB Diversity & 0.120 & 0.034 & 3.543 & \textless 0.001 & 0.054 & 0.187 \\
			Sharpness (less Blurriness) & -0.168 & 0.060 & -2.804 & 0.005 & -0.286 & -0.050 \\ \bottomrule
			\\

			Single line symbol \\ \midrule
			Edge density & 0.201 & 0.048 & 4.200 & \textless 0.001 & 0.107 & 0.296 \\
			RGB Diversity & 0.249 & 0.033 & 7.519 & \textless 0.001 & 0.184 & 0.314 \\
			Sharpness (less Blurriness) & -0.114 & 0.045 & -2.516 & 0.012 & -0.203 & -0.025 \\ \bottomrule
			\\
			\textbf{Dependent Variable: FN}\\
			\cmidrule{1-1}
			Double line symbol \\ \midrule
			Edge density & 0.371 & 0.066 & 5.650 & \textless 0.001 & 0.242 & 0.500 \\
			RGB Diversity & 0.081 & 0.035 & 2.323 & 0.020 & 0.013 & 0.149 \\
			Sharpness (less Blurriness) & -0.306 & 0.061 & -4.987 & \textless 0.001 & -0.427 & -0.186 \\ \bottomrule
			\\
			Single line symbol \\ \midrule
			Edge density & 0.299 & 0.049 & 6.082 & \textless 0.001 & 0.202 & 0.395 \\
			RGB Diversity & 0.094 & 0.034 & 2.787 & 0.005 & 0.028 & 0.161 \\
			Sharpness (less Blurriness) & -0.256 & 0.046 & -5.507 & \textless 0.001 & -0.347 & -0.165 \\ \bottomrule
			
	\end{tabular}}

	\label{table2}
\end{table}

Based on the results of the regression analysis, it can be concluded that edge density has a significant and positive effect on the number of FP errors in both symbol types. It means that as the clutter of a map tile increases, the number of boxes incorrectly categorized as road intersections grows too. This effect is the highest in the maps with double line symbol of the roads. As it was asserted in section 4.2 the main reason for this effect could be the fact that the existence of the flankers within an image full of the edges is higher compared with a more homogeneous one. The same effect of the edge density can be seen on the number of FN errors. One of the possible reasons is that as the crowdedness enlarges in an image, the probability that the intersections are covered or deformed by other objects increases too (Figures \ref{fig6}-c and \ref{fig7}-c). \cite{Volokitin2017} showed the same impact of image complexity on deep CNNs based object detection framework by asserting that the flankers existing in complex images could change the final accuracy dramatically.
The number of distinct RGB combinations has also a significant positive effect on FP and FN errors possibility in both symbol types. According to Table \ref{table2}, more unique RGB combinations cause more errors in the results. This factor shows a stronger effect on the number of FP than FN errors and also single line symbol than double line symbol of the roads. Although \cite{Grm2018} showed that deep CNNs are robust to changes in the input color space (unlike the traditional computer vision algorithms \citep{Kanan2012}), here it can be noticed that the variation within RGB space can be responsible for the number of errors even more than other factors in some cases such as FP errors of the maps with single line road symbols.
Finally, image sharpness shows a significant negative impact on both FP and FN errors in both map types. In other words, the blurrier a map tile was, the more of its objects were categorized incorrectly. The main reason of this effect could be the fact that blur removes the textures of the images \citep{Dodge2016}. This factor shows a stronger impact on the number of FN than FP errors and also double line symbol than single line symbol of the roads. Although the data augmentation approaches used in this study are a practical way to reduce the blurriness effect \citep{Grm2018}, this effect is still an important problem in the deep CNNs based object detection methods \citep{Dodge2016,Karahan2016}.
\section{Conclusion and future work}
The object detection frameworks such as Faster RCNN that are based on deep CNNs are easy to implement and have shown more accuracy than the traditional methods in many fields such as remote sensing which makes them more reliable and robust compared with traditional computer vision algorithms. Thus, in this paper, we used Faster RCNN framework to extract the road intersections from the scanned historical maps of USGS. We implemented the model for the maps with single line symbol and double line symbol of the roads separately for the sake of comparison. We have found that both models showed higher F1-scores compare to all traditional computer vision algorithms except for one case of single line symbols which needs a prior knowledge about the color codes which exist in the maps. 
We have also shown that the amount of errors in the outputs of Faster RCCN models is sensitive to tile complexity and blurriness as well as the number of distinct RGB combinations within it. Clutter in a map tile causes more FP errors because of the high number of flankers and also causes more FN errors because of more possibility of covering and deforming the road intersections by other objects. More RGB combinations and blurriness also increase the errors.

We used the version of deep CNN which has been trained on Microsoft COCO dataset consisting of 328k images. One possible extension to our study is using the deep CNNs which are pre trained by geospatial data such as remote sensing images to see if it can reduce the training time even more and help to improve the accuracy. Recently, it has been claimed that the essence of the structures of the objects in 2D geospatial images is different from other types of images because of the huge variation in their scales, orientations and shapes \citep{Xia2017}.
Finally, since the road intersections are considered as point data, we did not need to make the exact segment of them in the final results. However, for a lot of detection tasks such as extracting the whole road network we need to obtain the exact borders of the road surfaces. This would be a difficult task because of the resemblance among objects and the inconsistency in the historical maps. Recently, several deep CNN based object detection frameworks with the ability of segmentation have been proposed such as Mask RCNN \citep{He2017}. These systems provide the exact border of the detected object in their outputs. Some studies in the geospatial fields tried to use deep CNNs with the ability of segmentation to extract the road network from satellite images \citep{Henry2018}. As an extension to our study, it is possible to use the capabilities of these systems to extract the whole road networks and consequently, their intersections from the scanned historical maps.

\section*{Acknowledgement(s)}

\section*{Disclosure statement}

\section*{Funding}

\section*{Notes on contributor(s)}
Mahmoud Saeedimoghaddam is a Ph.D. student at the geography \& GIS department of the university of Cincinnati. His research is focused on analyzing the spatial complex systems such as urban areas. He is also interested in using machine learning approach in addressing the spatial issues.

Tomasz Stepinski is the Thomas Jefferson Chair Professor of Space Exploration at the University of Cincinnati and a Director of Space Informatics Lab. His recent area of research is a development of automated tools for intelligent and intuitive exploration of very large Earth and planetary datasets. He led the team who developed the GeoPAT2 – a toolbox for pattern-based spatial analysis. He is also interested in computational approaches to geodemographics, racial segregation and diversity.

\bibliographystyle{tfv}
\bibliography{Paper}

\end{document}